\documentclass[a4paper]{article}

\usepackage{INTERSPEECH2016}

\usepackage{graphicx}
\usepackage{amssymb,amsmath,bm}
\usepackage{textcomp}
\usepackage{multirow}
\usepackage{subfig}
\usepackage{balance}
\usepackage{url}

\def\RR{\mathbb R}

 \ninept
\title{Single-Channel Multi-Speaker Separation using Deep Clustering}

\makeatletter
\def\name#1{\gdef\@name{#1\\}}
\makeatother
\name{{\em Yusuf Isik$^{1,2,3}$\thanks{\noindent This work was done while Y.\ Isik was an intern at MERL.}, Jonathan Le Roux$^1$, Zhuo Chen$^{1,4}$,}\\
      {\em Shinji Watanabe$^1$, John R. Hershey$^1$}}

\address{$^1$ Mitsubishi Electric Research Laboratories (MERL), Cambridge, MA, USA \\
  $^2$ Sabanci University, Istanbul, Turkey, \quad
  $^3$ TUBITAK BILGEM, Kocaeli, Turkey \\
  $^4$ Columbia University, New York, NY, USA \\
  }

\begin{document}

\maketitle
\begin{abstract}
Deep clustering is a recently introduced deep learning architecture that uses discriminatively trained embeddings as the basis for clustering.  It was recently applied to spectrogram segmentation, resulting in impressive results on speaker-independent multi-speaker separation.  In this paper we extend the baseline system with an end-to-end signal approximation objective that greatly improves performance on a challenging speech separation.  We first significantly improve upon the baseline system performance by incorporating better regularization, larger temporal context, and a deeper architecture, culminating in an overall improvement in signal to distortion ratio (SDR) of 10.3 dB compared to the baseline of 6.0 dB for two-speaker separation, as well as a 7.1 dB SDR improvement for three-speaker separation.  We then extend the model to incorporate an enhancement layer to refine the signal estimates, and perform end-to-end training through both the clustering and enhancement stages to maximize signal fidelity.  We evaluate the results using automatic speech recognition.  The new signal approximation objective, combined with end-to-end training,  produces unprecedented performance, reducing the word error rate (WER) from 89.1\% down to 30.8\%.    This  represents a major advancement towards solving the cocktail party problem.  
\end{abstract}

\section{Introduction}

The human auditory system gives us the extraordinary ability to converse in the midst of a noisy throng of party goers. Solving this so-called \emph{cocktail party} problem \cite{bregman1990auditory} has proven extremely challenging for computers, and separating and recognizing speech in such conditions has been the holy grail of speech processing for more than 50 years. Previously, no practical method existed that could solve the problem in general conditions, especially in the case of single channel speech mixtures.

This work builds upon recent advances in single-channel separation, using a method known as \emph{deep clustering} \cite{Hershey2016ICASSP03}.  In deep clustering, a neural network is trained to assign an embedding vector to each element of a multi-dimensional signal, such that clustering the embeddings yields a desired segmentation of the signal.  In the cocktail-party problem, the embeddings are assigned to each time-frequency (TF) index of the short-time Fourier transform (STFT) of the mixture of speech signals.   Clustering these embeddings yields an assignment of each TF bin to one of the inferred sources.  These assignments are used as a masking function to extract the dominant parts of each source.  Preliminary work on this method produced remarkable performance, improving SNR by 6 dB on the task of separating two unknown speakers from a single-channel mixture \cite{Hershey2016ICASSP03}.  

In this paper we present improvements and extensions that enable a leap forward in separation quality, reaching levels of improvement  that were previously out of reach (audio examples and scripts to generate the data used here are available at \cite{dpcl_demos}).
In addition to improvements to the training procedure, we investigate the three speaker case, showing generalization between two- and three-speaker networks.  

The original deep clustering system was intended to only recover a binary masks for each source, leaving recovery of the missing features to subsequent stages.  In this paper, 
we incorporate enhancement layers to refine the signal estimate.   Using soft clustering,   we can then train the entire system \emph{end-to-end}, training jointly through the deep clustering embeddings, the clustering and enhancement stages.   This allows us to directly use a signal approximation objective instead of the original mask-based deep clustering objective.
 
Prior work in this area includes auditory grouping approaches to \emph{computational auditory scene analysis} (CASA) \cite{Cooke91,hu2013unsupervised}. These methods used hand-designed features to cluster the parts of the spectrum belonging to the same source.  Their success was limited by the lack of a machine learning framework.  Such a framework was provided in subsequent work on spectral clustering \cite{bach2006learning}, at the cost of prohibitive  complexity. 

Generative models have also been proposed, beginning with \cite{Ellis96}.  In constrained tasks, super-human speech separation was first achieved using factorial HMMs \cite{hershey2010super,rennie2010single,cooke2010monaural,Virtanen06} and was extended to speaker-independent separation \cite{weiss2009underdetermined}.   Variants of non-negative matrix factorization \cite{smaragdis2007convolutive, simsekli_non-negative_2014} and Bayesian non-parametric models \cite{blei2010bayesian,nakano2011bayesian} have also been used.
These methods suffer from  computational complexity and difficulty of discriminative training.

In contrast to the above,  deep learning approaches have recently provided fast and accurate methods on simpler enhancement tasks \cite{weninger2015speech, wang2014training, xu2014experimental, huang2015joint}. These methods treat the mask inferance as a classification problem, and hence can be discriminatively trained for  accuracy without sacrificing speed.   However they fail to learn in the speaker-independent case, where sources are of the same class \cite{Hershey2016ICASSP03}, despite the work-around of choosing the best permutation of the network outputs during training.  
We call this the \emph{permutation problem}:  there are multiple valid output masks that differ only by a permutation of the order of the sources, so a global decision is needed to choose a permutation.  

Deep clustering solves the permutation problem by framing mask estimation as a clustering problem.   To do so, it produces an embedding for each time-frequency element in the spectrogram, such that clustering the embeddings produces the desired segmentation.  The representation is thus independent of permutation of the source labels. It can also flexibly represent any number of sources, allowing the number of inferred sources to be decided at test time.  Below we present the deep clustering model and further investigate its capabilities.  We then present extensions to allow end-to-end training for signal fidelity.  

The results are evaluated using an automatic speech recognition model trained on clean speech.  The end-to-end signal approximation produces unprecedented performance, reducing the word error rate (WER) from close to 89.1\% WER down to 30.8\% by using the end-to-end training.  This  represents a major advancement towards solving the cocktail party problem.

\section{Deep Clustering Model}

Here we review the deep clustering formalism presented in \cite{Hershey2016ICASSP03}.
We define as $x$ a raw input signal and as $X_i = g_i(x), i \in \{1,\dots,N\},$ a feature vector indexed by an element $i$. In audio signals, $i$ is typically a TF index $(t,f)$, where $t$ indexes frame of the signal, $f$ indexes frequency, and $X_i = X_{t,f}$ the value of the complex spectrogram at the corresponding TF bin.
We assume that the TF bins can be partitioned into 
sets of TF bins in which each source dominates.
Once estimated, the partition for each source serves as a TF mask to be applied to $X_i$, yielding the TF components of each source that are uncorrupted by other sources.  The STFT can then be inverted to obtain estimates of each isolated source. The target partition in a given mixture is represented by the indicator $Y = \{y_{i,c}\}$,  mapping each element $i$ to each of $C$  components of the mixture, so that $y_{i,c} = 1$ if element $i$ is in cluster $c$.  Then $A=YY^T$ is a binary affinity matrix that represents the cluster assignments in a permutation-independent way:  $A_{i,j} = 1$ if $i$ and $j$ belong to the same cluster and $A_{i,j} = 0$ otherwise, and  $(YP)(YP)^T = YY^T$ for any permutation matrix $P$. 

To estimate the partition, we seek $D$-dimensional embeddings $V = f_{\theta}(x) \in \RR^{N \times D}$, parameterized by $\theta$, such that clustering the embeddings yields a partition of $\{1,\dots,N\}$ that is close to the target. 
In \cite{Hershey2016ICASSP03} and this work, $V = f_{\theta}(X)$ is based on a deep neural network that is a global function of the entire input signal $X$.
Each embedding $v_i \in \RR^D$ has unit norm, i.e., $|v_i|^2=1$.
We consider the embeddings $V$ to implicitly represent an  $N \times N$  estimated affinity matrix  $\hat{A}=VV^T$, and we optimize the embeddings such that, for an input $X$, $\hat{A}$ matches the ideal affinities $A$. This is done by minimizing, with respect to $V=f_{\theta}(X)$, the training cost function 
\begin{equation}
\label{eqn:deep_clustering_cost}
\mathcal{C}_{Y}(V) = \|  \hat{A} - A \|_{\mathrm{F}}^2 =\|  VV^{T} - YY^{T} \|_{\mathrm{F}}^2
\end{equation}
summed over training examples,  where $\|\cdot \|_{\mathrm{F}}^2$ is the squared Frobenius norm. 
Due to its low-rank nature, the objective and its gradient can be formulated so as to avoid operations on all pairs of elements, leading to an efficient implementation.

At test time, the embeddings $V = f_{\theta}(X)$ are computed on the test signal $X$, and the rows $v_i \in \RR^D$ are clustered using $K$-means. The resulting cluster assignments $\bar{Y}$ are used as binary masks on the complex spectrogram of the mixture, to estimate  the sources.  

\section{Improvements to the Training Recipe}

We investigated several approaches to improve performance over the baseline deep clustering method, including regularization such as drop-out, model size and shape, and training schedule.   We used the same feature extraction procedure as in \cite{Hershey2016ICASSP03}, with  log-magnitude STFT features as input, and we performed global mean-variance normalization as a pre-processing step.  For all experiments we used we used rmsprop optimization \cite{Tieleman2012}  with a fixed learning rate schedule, and early stopping based on cross-validation.

{\bf Regularizing recurrent network units:} Recurrent neural network (RNN) units, in particular LSTM structures, have been widely adopted in many tasks such as object detection, natural language processing, machine translation, and speech recognition.   Here we experiment with regularizing them using dropout. 

LSTM nodes consist of a recurrent memory cell surrounded by gates controlling its input, output, and recurrent connections.  The direct recurrent connections are element-wise and linear with weight 1, so that with the right setting of the gates, the memory is  perpetuated, and otherwise more general recurrent processing is obtained.

Dropout is a training regularization in which nodes are randomly set to zero.  In recurrent network there is a concern that dropout could interfere with LSTM's memorization ability;  for example,  \cite{zaremba2014recurrent} used it only on feed-forward connections, but not on the recurrent ones.  \emph{Recurrent dropout} samples the set of dropout nodes once for each sequence, and applies dropout to the same nodes at every time step for that sequence. Applying recurrent dropout to the LSTM memory cells recently yielded performance improvements on phoneme and speech recognition tasks with BLSTM acoustic models  \cite{moon2015rnndrop}.

In this work, we sampled the dropout masks once at each time step for the forward connections, and only once for each sequence for the recurrent connections. We used the same recurrent dropout mask for each gate. 

{\bf Architecture:} We investigated using deeper and wider architectures. The neural network model used in \cite{Hershey2016ICASSP03} was a two layer bidirectional long short-term memory (BLSTM) network followed by a feed-forward layer to produce embeddings.  We show that expanding the network size improves performance for our task.  

{\bf Temporal context:}  During training, the utterances are divided into fixed length non-overlapping segments, and gradients are computed using shuffled mini-batches of these segments, as in \cite{Hershey2016ICASSP03}.  Shorter segments increase the diversity within each batch, and may make an easier starting point for training, since the speech does not change as much over the segment.  
However, at test time, the network and clustering are given the entire utterance, so that the permutation problem can be solved globally.  So we may also expect that training on longer segments would improve performance in the end.    

In experiments below, we investigate training segment lengths of 100 versus 400, and show that although the longer segments work better, pretraining with shorter segments followed by training with longer segments leads to better performance on this task.   This is an example of \emph{curriculum learning} \cite{bengio2009curriculum}, in which starting with an easier task improves learning and generalization.

{\bf Multi-speaker training:}
Previous experiments \cite{Hershey2016ICASSP03} showed preliminary results on generalization from two speaker training to a three-speaker separation task.   
Here we further investigate generalization from three-speaker training to two-speaker separation, as well as multi-style training on both two and three-speaker mixtures, and show that the multi-style training can achieve the best performance on both tasks.    

\section{Optimizing Signal Reconstruction}

Deep clustering solves the difficult problem of segmenting the spectrogram into regions dominated by each source. It does not however solve the problem of recovering the sources in regions strongly dominated by other sources. Given the segmentation, this is arguably an easier problem.
We propose to use a second-stage enhancement network to obtain better source estimates, in particular for the missing regions. 
For each source $c$, the enhancement network first processes the concatenation of the amplitude spectrogram $x$ of the mixture and that $\hat{s}_c$ of the deep clustering estimate through a BLSTM layer and a feed-forward linear layer, to produce an output $z_c$. Sequence-level mean and variance normalization is applied to the input, and the network parameters are shared for all sources. A soft-max is then used to combine the outputs $z_{c}$ across sources, forming a mask $m_{c,i}= {\mathrm{e}^{z_{c,i}}} / {\sum_{c'} \mathrm{e}^{z_{c',i}}}$ at each TF bin $i$. This mask is applied to the mixture, yielding the final estimate $\tilde{s}_{c,i} = m_{c,i} x_i$.
During training, we optimize the enhancement cost function $\mathcal{C}_{E}= \min_{\pi\in\mathcal{P}} \sum_{c,i} (s_{c,i} -\tilde{s}_{\pi(c),i})^2,$
where $\mathcal{P}$ is the set of permutations on $\{1,\dots,C\}$.  Since the enhancement network is trained to directly improve the signal reconstruction, it may improve upon deep clustering, especially in regions where the signal is dominated by other sources. 

\section{End-to-End Training}
In order to consider end-to-end training in the sense of jointly training the deep clustering with the enhancement stage, we need to compute gradients of the clustering step.  In \cite{Hershey2016ICASSP03}, hard $K$-means clustering was used to cluster the embeddings.  
The resulting binary masks cannot be directly optimized to improve signal fidelity, because the optimal masks are generally continuous, and because the hard clustering is not differentiable.   
Here we propose a soft $K$-means algorithm that enables us to directly optimize the estimated speech for signal fidelity.

In \cite{Hershey2016ICASSP03}, clustering was performed with equal weights on the TF embeddings, although weights were used in the training objective in order to train only on TF elements with significant energy.  Here we introduce similar weights  weights $w_i$ for each  embedding $v_i$ to focus the clustering on TF elements with significant energy.  The goal is mainly to avoid clustering silence regions, which may have noisy embeddings, and for which mask estimation errors are inconsequential.  

The soft weighted $K$-means algorithm can be interpreted as a weighted expectation maximization (EM) algorithm for a Gaussian mixture model with tied circular covariances.  It alternates between computing the assignment of every embedding to each centroid, and updating the centroids:
\vspace{-2mm}
\begin{align}
\gamma_{i,c} &= \frac{\mathrm{e}^{-\alpha |v_i - \mu_c|^2}}{\sum_{c'} \mathrm{e}^{-\alpha |v_i - \mu_{c'}|^2} },   \quad &
 \mu_{c} &= \frac{\sum_i \gamma_{i,c} w_i v_{i}}{\sum_i \gamma_{i,c} w_i},
 \label{eq:softkmeans}
\end{align}
where $\mu_{c}$ is the estimated mean of cluster $c$, and  $\gamma_{i,j}$ is the estimated assignment of embedding $i$ to the cluster $c$.  The parameter $\alpha$ controls the hardness of the clustering. As the value of $\alpha$ increases, the algorithm approaches $K$-means. 

The weights $w_i$ may be set in a variety of ways.    A reasonable choice could be to set $w_i$ according to the power of the mixture in each TF bin.  Here we set the weights to $1$, except in silence TF bins where the weight is set to $0$. Silence is defined using a threshold on the energy relative to the maximum of the mixture. 

End-to-end training is performed by \emph{unfolding} the steps of \eqref{eq:softkmeans}, and treating them as layers in a clustering network, according to the general framework known as \emph{deep unfolding} \cite{Hershey2014arXiv09}.   The gradients of each step are thus passed to the previous layers using standard back-propagation.  

\section{Experiments}

{\bf Experimental setup:}
We evaluate deep clustering on a single-channel speaker-independent speech separation task, considering mixtures of two and three speakers with all gender combinations. For two-speaker experiments, we use the corpus introduced in \cite{Hershey2016ICASSP03}, derived from the Wall Street Journal (WSJ0) corpus. It consists in a 30 h training set and a 10 h validation set with two-speaker mixtures  generated by randomly selecting utterances by different speakers from the WSJ0 training set {\footnotesize \verb|si_tr_s|}, and mixing them at various signal-to-noise ratios (SNR) randomly chosen between 0~dB and 10~dB.
The validation set was here used to optimize some tuning parameters. The 5 h test set consists in mixtures similarly generated using utterances from 16 speakers from the WSJ0 development set {\footnotesize \verb|si_dt_05|} and evaluation set {\footnotesize \verb|si_et_05|}. The speakers are different from those in our training and validation sets, leading to a speaker-independent separation task. 
For three-speaker experiments, we created a corpus similar to the two-speaker one, with the same amounts of data generated from the same datasets.
All data were downsampled to 8~kHz before processing to reduce computational and memory costs.
The input features $X$ were the log spectral magnitudes of the speech mixture, computed using a short-time Fourier transform (STFT) with a 32~ms sine window and 8~ms shift. 

The scores are reported in terms of signal-to-distortion ratio (SDR), which we define as scale-invariant SNR. As oracle upper bounds on performance for our datasets, we report in Table~\ref{tab:ideal_results} the results obtained using two types of ``ideal'' masks: the ideal binary mask (ibm) defined as $ a^{\mathrm{ibm}}_i = \delta(|s_i| > \max_{j\neq i}|s_j|) $, which leads to highest SNR among all binary masks, and a ``Wiener-like'' filter (wf) defined as $ a^{\mathrm{wf}}_i = |s_i|^2 / \sum_j|s_j|^2$, which empirically leads to good SNR,  with values in $[0,1]$ \cite{wang2005ideal,Erdogan2015ICASSP04}. Here $s_i$ denotes the time-frequency representation of speaker $i$. CASA \cite{hu2013unsupervised} and previous deep clustering  \cite{Hershey2016ICASSP03} results are also shown for the two-speaker set.

\begin{table}[h]
\vspace{-0.1cm}
\caption{SDR (dB) improvements using the ideal binary mask (ibm), oracle Wiener-like filter (wf), compared to prior methods dpcl \cite{Hershey2016ICASSP03} and CASA \cite{hu2013unsupervised} on the two- and three-speaker test sets.}
\vspace{-0.3cm}
\label{tab:ideal_results}
\begin{center}
{\footnotesize
\begin{tabular}{lcccc}
\hline 
\# speakers & ibm & wf & dpcl v1 \cite{Hershey2016ICASSP03} & CASA \cite{hu2013unsupervised}\\
\hline
$2$  & 13.5 & 13.9 & 6.0 & 3.1 \\
$3$  & 13.3 & 13.8 & - & - \\
\hline 
\end{tabular}
}
\end{center}
\vspace{-0.7cm}
\end{table}

The initial system,  based on \cite{Hershey2016ICASSP03},  trains a deep clustering model on 100-frame segments from the two-speaker mixtures. The network, with 2 BLSTM layers, each having 300 forward and 300 backward LSTM cells, is denoted as $300\!\times\!2$. The learning rate for the rmsprop algorithm \cite{Tieleman2012} was  $\lambda = 0.001 \times (1/2)^{\lfloor\epsilon/50\rfloor}$, where $\epsilon$ is the epoch number.  

{\bf Regularization:}   We first considered improving performance of the baseline using common regularization practices.   Table \ref{tab:sdr_decomposition} shows the contribution of dropout ($p = 0.5$) on feed-forward connections, recurrent dropout ($p=0.2$), and gradient normalization ($|\nabla|\leq 200$), where the parameters were tuned on development data.   Together these result in a 3.3 dB improvement in SDR relative to the baseline.

\begin{table}[h!] \vspace{-0.2cm}
\caption{Decomposition of the SDR improvements (dB) on the two-speaker test set using $300\!\times\!2$ model.}
\vspace{-0.2cm}
\label{tab:sdr_decomposition}
\begin{center}
{\footnotesize \setlength{\tabcolsep}{5pt}
\begin{tabular}{cccc}
\hline 
rmsprop & +dropout & +recurrent dropout & +norm constraint \\
\hline
5.7  & 8.0 & 8.9 & 9.0\\
\hline 
\end{tabular}
}
\end{center}
\vspace{-0.6cm}
\end{table}

{\bf Architecture:} Various network architectures were investigated by increasing the number of hidden units and number of BLSTM layers, as shown in Table \ref{tab:T100_results}. An improvement of 9.4~dB SDR was obtained with a deeper $300 \times 4$ architecture, with 4 BLSTM layers and 300 units in each LSTM.

\vspace{-0.2cm}
\begin{table}[!h]
\caption{SDR (dB) improvements on the two-speaker test set for different architecture sizes.} 
\vspace{-0.2cm}
\label{tab:T100_results}
\begin{center}
{\footnotesize
\begin{tabular}{lccc}
\hline 
model & same-gender & different-gender & overall \\
\hline
$300\!\times\!2$  & 6.4 & 11.2 & 9.0 \\
$600\!\times\!2$  & 6.1 & 11.5 & 9.0 \\
$300\!\times\!4$  & 7.1 & 11.5 & 9.4 \\
\hline 
\end{tabular}
}
\end{center}
\vspace{-0.7cm}
\end{table}

{\bf Pre-training of temporal context:} 
Training the model with segments of 400 frames, after pre-training using 100-frame segments,  boosts performance to 10.3 dB, as shown in Table \ref{tab:T400_results}, from $9.9$~dB without pre-training. Results for the remaining experiments are based on the pre-trained $300\!\times\!4$ model.  

\vspace{-0.2cm}
\begin{table}[!h]
\caption{SDR (dB) improvements on the two-speaker test set after training with 400 frame length segments.}
\vspace{-0.2cm}
\label{tab:T400_results}
\begin{center}
{\footnotesize
\begin{tabular}{lccc}
\hline 
model & same-gender & different-gender & overall \\
\hline
$600\!\times\!2$  & 7.8 & 11.7 & \phantom{0}9.9 \\
$300\!\times\!4$  & 8.6 & 11.7 & 10.3 \\
\hline 
\end{tabular}
}
\end{center}
\vspace{-0.7cm}
\end{table}

\noindent {\bf Multi-speaker training:} We train the model further with a blend of two- and three-speaker mixtures. For comparison, we also trained a model using only three-speaker mixtures, again training first over 100-frame segments, then over 400-frame segments. The performance of the models trained on two-speaker mixtures only, on three-speaker mixtures only, and using the multi-speaker training, are shown in Table~\ref{tab:multi_speaker}. The three-speaker mixture model seems to generalize better to two speakers than vice versa, whereas the multi-speaker trained model performed the best on both tasks.

\begin{table}[!h]
\vspace{-.2cm}
\caption{Generalization across different numbers of speakers in terms of SDR improvements (dB).}
\vspace{-0.2cm}
\label{tab:multi_speaker}
\begin{center}
{\footnotesize \setlength{\tabcolsep}{15pt}
\begin{tabular}{lcc}
\hline 
\multirow{2}{*}{\bf Training data} &\multicolumn{2}{c}{\bf Test data } \\
 & $2$ speaker  & $3$ speaker \\ 
\hline      
$2$ speaker      &  10.3 & 2.1 \\
$3$ speaker      &  \phantom{0}8.5  & 7.1 \\
Mixed curriculum &  10.5  & 7.1 \\
\hline 
\end{tabular}
}
\end{center}
\vspace{-0.7cm}
\end{table}

{\bf Soft clustering:} The choice of the clustering hardness parameter $\alpha$ and the weights on TF bins is analyzed on the validation set, with results in Table \ref{tab:kmeans_results}. The use of weights to ignore silence improves performance with diminishing returns for larg $\alpha$.  The best result is for $\alpha=5$.

\begin{table}[!h]
\vspace{-.2cm}
\caption{Performance as a function of soft weighted $K$-means parameters on the two-speaker validation set.}
\vspace{-0.2cm}
\label{tab:kmeans_results}
\begin{center}
{\footnotesize
\begin{tabular}{lcccc}
\hline 
weights & $\alpha=2$ & $\alpha=5$ & $\alpha=10$ & hard $K$-means  \\
\hline
all equal    & 5.0  & 10.1 & 10.1 & 10.3 \\
mask silent  & 9.1  & 10.3 & 10.2 & 10.3  \\
\hline 
\end{tabular}
}
\end{center}
\vspace{-0.7cm}
\end{table}

\noindent {\bf End-to-end training:} Finally, we investigate end-to-end training, using a second-stage enhancement network on top of the deep clustering (`dpcl') model. Our enhancement network features two BLSTM layers with 300 units in each LSTM layer, with one instance per source followed by a soft-max layer to form a masking function. We first trained the enhancement network separately (`dpcl + enh'), followed by end-to-end fine-tuning in combination with the dpcl model (`end-to-end'). Table \ref{tab:enhancement_results} shows the improvement in SDR as well as \emph{magnitude SNR} (SNR computed on the magnitude spectrograms).

\begin{table}[!h]
\vspace{-.2cm}
\caption{SDR / Magnitude SNR improvements (dB) and WER  with enhancement network.}
\vspace{-0.2cm}
\label{tab:enhancement_results}
\begin{center}
{\footnotesize \setlength{\tabcolsep}{3pt}
\begin{tabular}{lccc|c}
\hline 
model & same-gender & different-gender & overall & WER \\
\hline
dpcl           & 8.6 / \phantom{0}8.9 & 11.7 / 11.4 & 10.3 / 10.2 & 87.9 \% \\
dpcl + enh & 9.1 / 10.7 & 11.9 / 13.6 & 10.6 / 12.3 & 32.8 \% \\
end-to-end & 9.4 / 11.1  & 12.0 / 13.7 & 10.8 / 12.5 & 30.8\% \\
\hline 
\end{tabular}
}
\end{center}
\vspace{-0.7cm}
\end{table}

\begin{figure}[h!]
	\centering
    		\includegraphics[width=0.85\columnwidth]{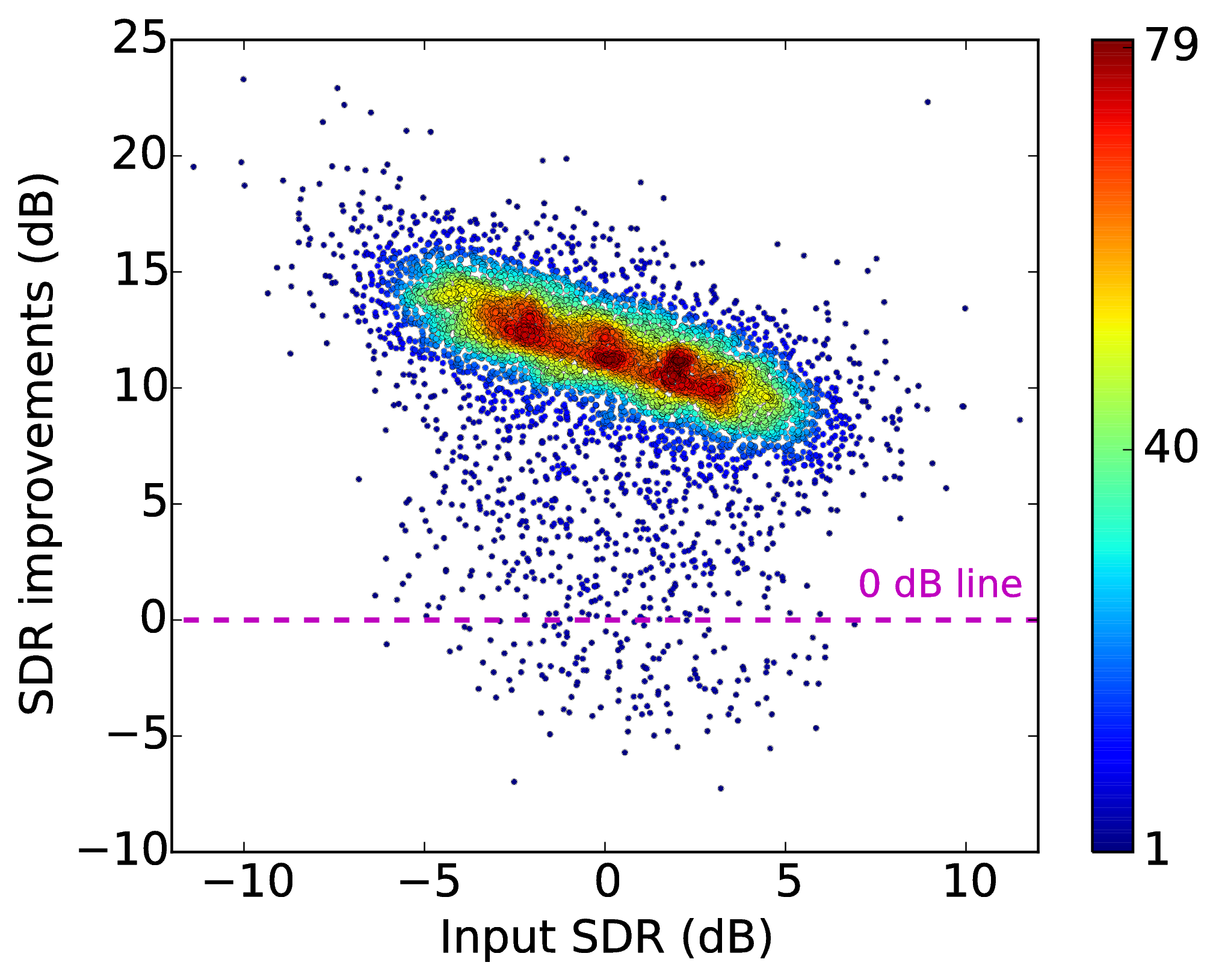}
				\caption{Scatter plot for the input SDRs and the corresponding improvements. Color indicates density.} 	\label{fig:scatter}	
	\vspace{-.3cm}
\end{figure}

The magnitude SNR is insensitive to phase estimation errors introduced by using the noisy phases for reconstruction, whereas the SDR might get worse as a result of phase errors, even if the amplitudes are accurate.
Speech recognition uses features based on the amplitudes, and hence the improvements in magnitude SNR seem to predict the improvements in WER due to the enhancement and end-to-end training. 
Fig.~\ref{fig:scatter} shows that the SDR improvements of the end-to-end model are consistently good on nearly all of the two-speaker test mixtures.

\noindent {\bf ASR performance:} We evaluated ASR performance (WER) with GMM-based clean-speech WSJ models obtained by a standard Kaldi recipe \cite{povey2011kaldi}. The noisy baseline result on the mixtures is 89.1 \%, while the result on the clean speech is 19.9 \%.
The raw output from dpcl did not work well, despite good perceptual quality, possibly due to the effect of near-zero values in the masked spectrum, which is known to degrade ASR performance.
However, the enhancement networks significantly mitigated the degradation, and finally obtained 30.8 \% with the end-to-end network.

\begin{figure}[h!]
	\centering    
	    \subfloat[onset]{
	\includegraphics[width=0.22\columnwidth]{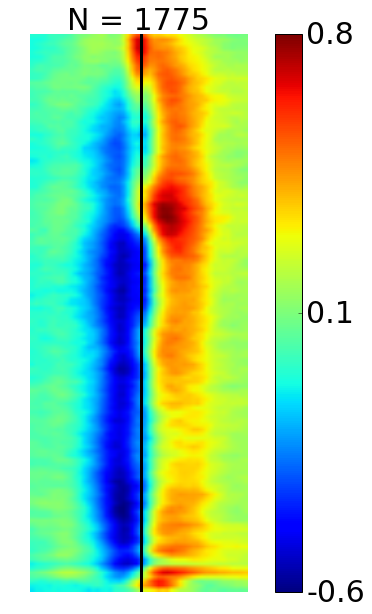}
	}\label{fig:revcor:onset}
	\subfloat[pitch]{
	\includegraphics[width=0.22\columnwidth]{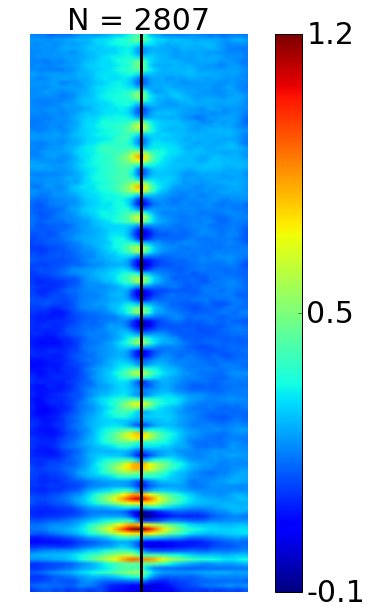}
	}\label{fig:revcor:pitch}
    \subfloat[chirp]{
	\includegraphics[width=0.22\columnwidth]{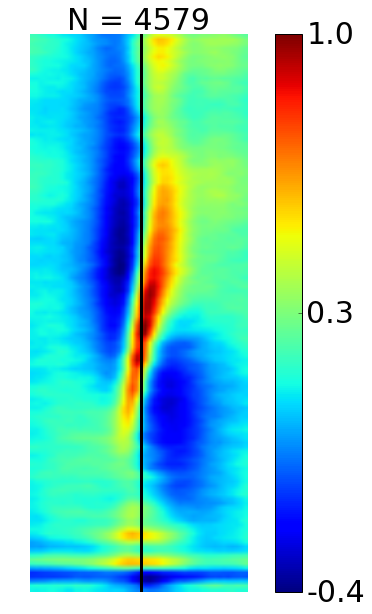}
	}\label{fig:revcor:chirp}
    \subfloat[transition]{
	\includegraphics[width=0.22\columnwidth]{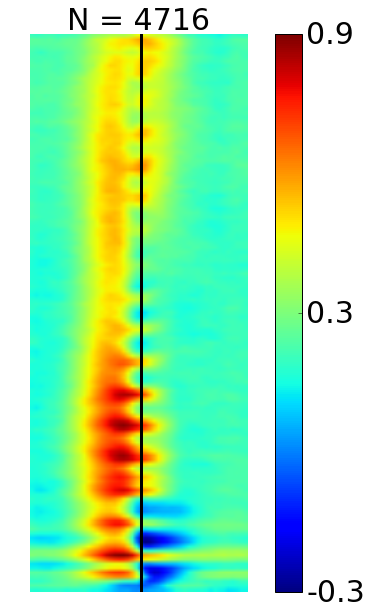}
	}\label{fig:revcor:transition}
 			\caption{Example spike-triggered spectrogram averages with 50-frame context, for active LSTM nodes in the second layer. $N$ is the number of active frames for the corresponding node.}
		\label{fig:revcor}
\end{figure}

\noindent {\bf Visualization:} To gain insight into network functioning, we performed reverse correlation experiments.  For each node, we average the 50-frame patches of input centered at the time  when the node is active (e.g., the node is at 80\% of its maximum value).  
Fig.~\ref{fig:revcor} shows a variety of interesting patterns, which seem to reflect such properties as onsets,  pitch,  frequency chirps, and vowel-fricative transitions.   

{\bf Conclusion:} We have improved and extended the deep clustering framework to perform end-to-end training for signal reconstruction quality for the first time.   We show significant improvements to performance both on signal quality metrics and speech recognition error rates.

\newpage
\balance
\bibliographystyle{IEEEtran}


\end{document}